# Deep Video Portraits

HYEONGWOO KIM, Max Planck Institute for Informatics, Germany
PABLO GARRIDO, Technicolor, France
AYUSH TEWARI and WEIPENG XU, Max Planck Institute for Informatics, Germany
JUSTUS THIES and MATTHIAS NIESSNER, Technical University of Munich, Germany
PATRICK PÉREZ, Technicolor, France
CHRISTIAN RICHARDT, University of Bath, United Kingdom
MICHAEL ZOLLHÖFER, Stanford University, United States of America
CHRISTIAN THEOBALT, Max Planck Institute for Informatics, Germany

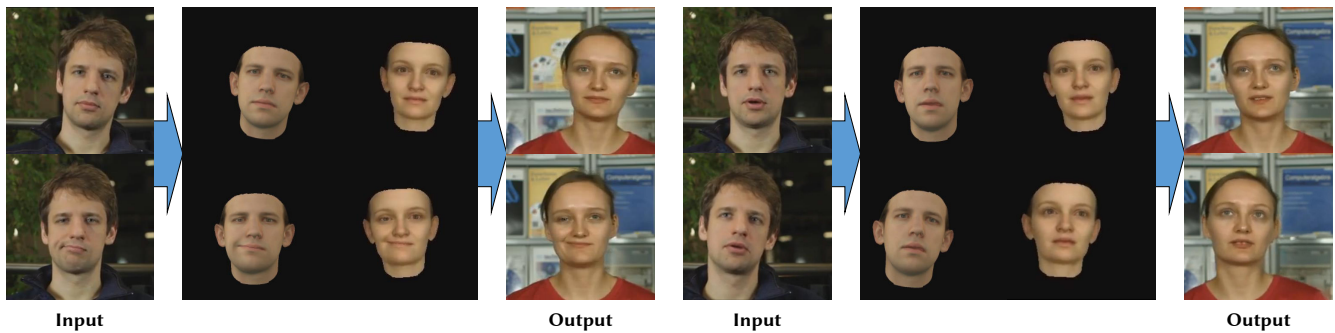

Fig. 1. Unlike current face reenactment approaches that only modify the expression of a target actor in a video, our novel deep video portrait approach enables full control over the target by transferring the rigid head pose, facial expression and eye motion with a high level of photorealism.

We present a novel approach that enables photo-realistic re-animation of portrait videos using only an input video. In contrast to existing approaches that are restricted to manipulations of facial expressions only, we are the first to transfer the full 3D head position, head rotation, face expression, eye gaze, and eye blinking from a source actor to a portrait video of a target actor. The core of our approach is a generative neural network with a novel space-time architecture. The network takes as input synthetic renderings of a parametric face model, based on which it predicts photo-realistic video frames for a given target actor. The realism in this rendering-to-video transfer is achieved by careful adversarial training, and as a result, we can create modified target videos that mimic the behavior of the synthetically-created input. In order to enable source-to-target video re-animation, we render a synthetic target video with the reconstructed head animation parameters from a source video, and feed it into the trained network – thus taking full control of the target. With the ability to freely recombine source and target parameters, we are able to demonstrate a large variety of video rewrite applications without explicitly modeling hair, body or background. For instance, we can reenact the full head using interactive user-controlled editing, and realize high-fidelity visual dubbing. To demonstrate the high quality of our output, we conduct an extensive series of experiments and evaluations, where for instance a user study shows that our video edits are hard to detect.

Authors' addresses: Hyeongwoo Kim, Max Planck Institute for Informatics, Campus E1.4, Saarbrücken, 66123, Germany, hyeongwoo.kim@mpi-inf.mpg.de; Pablo Garrido, Technicolor, 975 Avenue des Champs Blancs, Cesson-Sévigné, 35576, France, pablo.garrido.adrian@gmail.com; Ayush Tewari, atewari@mpi-inf.mpg.de; Weipeng Xu, wxu@mpi-inf.mpg.de, Max Planck Institute for Informatics, Campus E1.4, Saarbrücken, 66123, Germany; Justus Thies, justus.thies@tum.de; Matthias Nießner, niessner@tum.de, Technical University of Munich, Boltzmannstraße 3, Garching, 85748, Germany; Patrick Pérez, Technicolor, 975 Avenue des Champs Blancs, Cesson-Sévigné, 35576, France, Patrick.Perez@technicolor.com; Christian Richardt, University of Bath, Claverton Down, Bath, BA2 7AY, United Kingdom, christian@richardt.name; Michael Zollhöfer, Stanford University, 353 Serra Mall, Stanford, CA, 94305, United States of America, zollhoefer@cs.stanford.edu; Christian Theobalt, Max Planck Institute for Informatics, Campus E1.4, Saarbrücken, 66123, Germany, theobalt@mpi-inf.mpg.de.

## 1 INTRODUCTION

Synthesizing and editing video portraits, i.e., videos framed to show a person's head and upper body, is an important problem in computer graphics, with applications in video editing and movie post-production, visual effects, visual dubbing, virtual reality, and telepresence, among others. In this paper, we address the problem of synthesizing a photo-realistic video portrait of a *target* actor that mimics the actions of a *source* actor, where source and target can be different subjects. More specifically, our approach enables a source actor to take full control of the rigid head pose, face expressions and eye motion of the target actor; even face identity can be modified to some extent. All of these dimensions can be manipulated together or independently. Full target frames, including the entire head and hair, but also a realistic upper body and scene background complying with the modified head, are automatically synthesized.

Recently, many methods have been proposed for face-interior reenactment [Liu et al. 2001; Olszewski et al. 2017; Suwajanakorn et al. 2017; Thies et al. 2015, 2016; Vlasic et al. 2005]. Here, only the face expression can be modified realistically, but not the full 3D head pose, including a consistent upper body and a consistently changing background. Many of these methods fit a parametric 3D face model to RGB(-D) video [Thies et al. 2015, 2016; Vlasic et al. 2005], and re-render the modified model as a blended overlay over the target video for reenactment, even in real time [Thies et al. 2015, 2016]. Synthesizing a complete portrait video under full 3D head control is much more challenging. Averbuch-Elor et al. [2017] enable mild head pose changes driven by a source actor based on image warping. They generate reactive dynamic profile pictures from a static target portrait photo, but not fully reenacted videos. Also, large changes in head pose cause artifacts (see Section 7.3), the target gaze cannot be controlled, and the identity of the target person is not fully preserved (mouth appearance is copied from the source actor).

Performance-driven 3D head animation methods [Cao et al. 2015, 2014a, 2016; Hu et al. 2017; Ichim et al. 2015; Li et al. 2015; Olszewski et al. 2016; Weise et al. 2011] are related to our work, but have orthogonal methodology and application goals. They typically drive the full head pose of stylized 3D CG avatars based on visual source actor input, e.g., for games or stylized VR environments. Recently, Cao et al. [2016] proposed image-based 3D avatars with dynamic textures based on a real-time face tracker. However, their goal is full 3D animated head control and rendering, often intentionally in a stylized rather than a photo-realistic fashion.

We take a different approach that directly generates entire photo-realistic video portraits in front of general static backgrounds under full control of a target's head pose, facial expression, and eye motion. We formulate video portrait synthesis and reenactment as a rendering-to-video translation task. Input to our algorithm are synthetic renderings of *only* the coarse and fully-controllable 3D face interior model of a target actor and separately rendered eye gaze images, which can be robustly and efficiently obtained via a state-of-the-art model-based reconstruction technique. The input is automatically translated into full-frame photo-realistic video output showing the entire upper body and background. Since we only track the face, we cannot actively control the motion of the torso or hair, or control the background, but our rendering-to-video translation network is able to implicitly synthesize a plausible body and background (including some shadows and reflections) for a given head pose. This translation problem is tackled using a novel space-time encoder–decoder deep neural network, which is trained in an adversarial manner.

At the core of our approach is a conditional generative adversarial network (cGAN) [Isola et al. 2017], which is specifically tailored to video portrait synthesis. For temporal stability, we use a novel space-time network architecture that takes as input short sequences of conditioning input frames of head and eye gaze in a sliding window manner to synthesize each target video frame. Our target and scene-specific networks only require a few minutes of portrait video footage of a person for training. To the best of our knowledge, our approach is the first to synthesize full photo-realistic video portraits of a target person's upper body, including realistic clothing and hair, and consistent scene background, under full 3D control of the target's head. To summarize, we make the following technical contributions:

- A rendering-to-video translation network that transforms coarse face model renderings into full photo-realistic portrait video output.
- A novel space-time encoding as conditional input for temporally coherent video synthesis that represents face geometry, reflectance, and motion as well as eye gaze and eye blinks.
- A comprehensive evaluation on several applications to demonstrate the flexibility and effectiveness of our approach.

We demonstrate the potential and high quality of our method in many intriguing applications, ranging from face reenactment and visual dubbing for foreign language movies to user-guided interactive editing of portrait videos for movie postproduction. A comprehensive comparison to state-of-the-art methods and a user study confirm the high fidelity of our results.

## 2 RELATED WORK

We discuss related optimization and learning-based methods that aim at reconstructing, animating and re-writing faces in images and videos, and review relevant image-to-image translation work. For a comprehensive overview of current methods we refer to a recent state-of-the-art report on monocular 3D face reconstruction, tracking and applications [Zollhöfer et al. 2018].

*Monocular Face Reconstruction.* Face reconstruction methods aim to reconstruct 3D face models of shape and appearance from visual data. Optimization-based methods fit a 3D template model, mainly the inner face region, to single images [Blanz et al. 2004; Blanz and Vetter 1999], unstructured image collections [Kemelmacher-Shlizerman 2013; Kemelmacher-Shlizerman et al. 2011; Roth et al. 2017] or video [Cao et al. 2014b; Fyffe et al. 2014; Garrido et al. 2016; Ichim et al. 2015; Shi et al. 2014; Suwajanakorn et al. 2014; Thies et al. 2016; Wu et al. 2016]. Recently, Booth et al. [2018] proposed a large-scale parametric face model constructed from almost ten thousand 3D scans. Learning-based approaches leverage a large corpus of images or image patches to learn a regressor for predicting either 3D face shape and appearance [Richardson et al. 2016; Tewari et al. 2017; Tran et al. 2017], fine-scale skin details [Cao et al. 2015], or both [Richardson et al. 2017; Sela et al. 2017]. Deep neural networks have been shown to be quite robust for inferring the coarse 3D facial shape and appearance of the inner face region, even when trained on synthetic data [Richardson et al. 2016]. Tewari et al. [2017] showed that encoder–decoder architectures can be trained fully unsupervised on in-the-wild images by integrating physical image formation into the network. Richardson et al. [2017] trained an end-to-end regressor to recover facial geometry at a coarse and fine-scale level. Sela et al. [2017] use an encoder–decoder network to infer a detailed depth image and a dense correspondence map, which serve as a basis for non-rigidly deforming a template mesh.

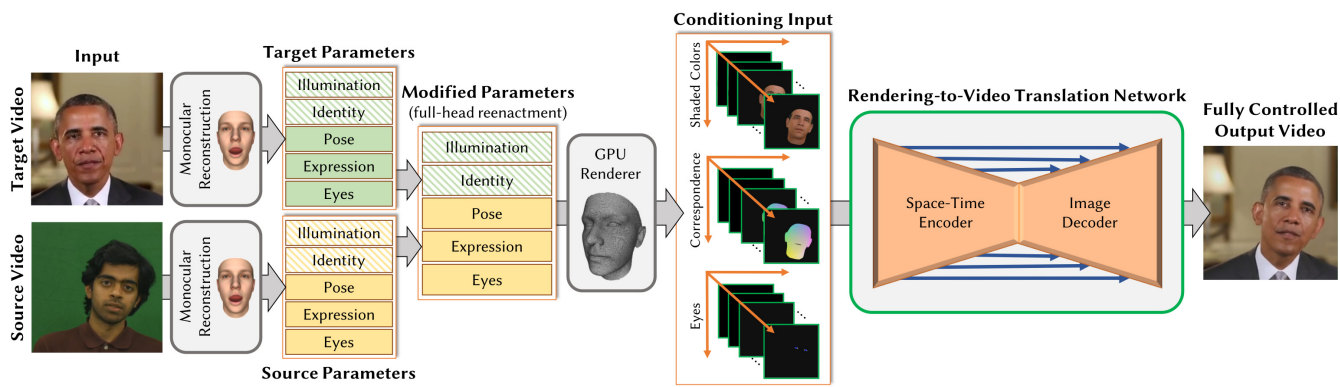

Fig. 2. Deep video portraits enable a source actor to fully control a target video portrait. First, a low-dimensional parametric representation (left) of both videos is obtained using monocular face reconstruction. The head pose, expression and eye gaze can now be transferred in parameter space (middle). We do not focus on the modification of the identity and scene illumination (hatched background), since we are interested in reenactment. Finally, we render conditioning input images that are converted to a photo-realistic video portrait of the target actor (right). *Obama* video courtesy of the White House (public domain).

Still, none of these methods creates a fully generative model for the entire head, hair, mouth interior, and eye gaze, like we do.

*Video-based Facial Reenactment.* Facial reenactment methods rewrite the face content of a target actor in a video or image by transferring facial expressions from a source actor. Facial expressions are commonly transferred via dense motion fields [Averbuch-Elor et al. 2017; Liu et al. 2001; Suwajanakorn et al. 2015], parameters [Thies et al. 2016, 2018; Vlasic et al. 2005], or by warping candidate frames that are selected based on the facial motion [Dale et al. 2011], appearance metrics [Kemelmacher-Shlizerman et al. 2010] or both [Garrido et al. 2014; Li et al. 2014]. The methods described above first reconstruct and track the source and target faces, which are represented as a set of sparse 2D landmarks or dense 3D models. Most approaches only modify the inner region of the face and thus are mainly intended for altering facial expressions, but they do not take full control of a video portrait in terms of rigid head pose, facial expression, and eye gaze. Recently, Wood et al. [2018] proposed an approach for eye gaze redirection based on a fitted parametric eye model. Their approach only provides control over the eye region.

One notable exception to pure facial reenactment is Averbuch-Elor et al.'s approach [2017], which enables the reenactment of a portrait image and allows for slight changes in head pose via image warping [Fried et al. 2016]. Since this approach is based on a single target image, it copies the mouth interior from the source to the target, thus preserving the target's identity only partially. We take advantage of learning from a target video to allow for larger changes in head pose, facial reenactment, and joint control of the eye gaze.

*Visual Dubbing.* Visual dubbing is a particular instance of face reenactment that aims to alter the mouth motion of the target actor to match a new audio track, commonly spoken in a foreign language by a dubbing actor. Here, we can find speech-driven [Bregler et al. 1997; Chang and Ezzat 2005; Ezzat et al. 2002; Liu and Ostermann 2011; Suwajanakorn et al. 2017] or performance-driven [Garrido et al. 2015; Thies et al. 2016] techniques. Speech-driven dubbing techniques learn a person-specific phoneme-to-viseme mapping from a training sequence of the actor. These methods produce accurate lip sync with visually imperceptible artifacts, as recently demonstrated by Suwajanakorn et al. [2017]. However, they cannot directly control the target's facial expressions. Performance-driven techniques overcome this limitation by transferring semantically-meaningful motion parameters and re-rendering the target model with photo-realistic reflectance [Thies et al. 2016], and fine-scale details [Garrido et al. 2015, 2016]. These approaches generalize better, but do not edit the head pose and still struggle to synthesize photo-realistic mouth deformations. In contrast, our approach learns to synthesize photo-realistic facial motion and actions from coarse renderings, thus enabling the synthesis of expressions and joint modification of the head pose, with consistent body and background.

*Image-to-image Translation.* Approaches using conditional GANs [Mirza and Osindero 2014], such as Isola et al.'s "pix2pix" [2017], have shown impressive results on image-to-image translation tasks which convert between images of two different domains, such as maps and satellite photos. These combine encoder–decoder architectures [Hinton and Salakhutdinov 2006], often with skip-connections [Ronneberger et al. 2015], with adversarial loss functions [Goodfellow et al. 2014; Radford et al. 2016]. Chen and Koltun [2017] were the first to demonstrate high-resolution results with 2 megapixel resolution, using cascaded refinement networks without adversarial training. The latest trends show that it is even possible to train high-resolution GANs [Karras et al. 2018] and conditional GANs [Wang et al. 2018] at similar resolutions. However, the main challenge is the requirement for paired training data, as corresponding image pairs are often not available. This problem is tackled by CycleGAN [Zhu et al. 2017], DualGAN [Yi et al. 2017], and UNIT [Liu et al. 2017] – multiple concurrent unsupervised image-to-image translation techniques that only require two sets of unpaired training samples. These techniques have captured the imagination of many people by translating between photographs and paintings, horses and zebras, face photos and depth as well as correspondence maps [Sela et al. 2017], and translation from face photos to cartoon drawings [Taigman et al. 2017]. Ganin et al. [2016] learn photo-realistic gaze manipulation in images. Olszewski et al. [2017] synthesize a realistic inner face texture, but cannot generate a fully controllable output video, including person-specific hair. Lassner et al. [2017] propose a generative model to synthesize people in clothing, and Ma et al. [2017] generate new images of persons in arbitrary poses using image-to-image translation. In contrast, our approach enables the synthesis of temporally-coherent video portraits that follow the animation of a source actor in terms of head pose, facial expression and eye gaze.

## 3 OVERVIEW

Our deep video portraits approach provides full control of the head of a *target actor* by transferring the rigid head pose, facial expression, and eye motion of a *source actor*, while preserving the target's identity and appearance. Full target video frames are synthesized, including consistent upper body posture, hair and background. First, we track the source and target actor using a state-of-the-art monocular face reconstruction approach that uses a parametric face and illumination model (see Section 4). The resulting sequence of low-dimensional parameter vectors represents the actor's identity, head pose, expression, eye gaze, and the scene lighting for every video frame (Figure 2, left). This allows us to transfer the head pose, expression, and/or eye gaze parameters from the source to the target, as desired. In the next step (Figure 2, middle), we generate new synthetic renderings of the target actor based on the modified parameters (see Section 5). In addition to a normal color rendering, we also render correspondence maps and eye gaze images. These renderings serve as conditioning input to our novel rendering-to-video translation network (see Section 6), which is trained to convert the synthetic input into photo-realistic output (see Figure 2, right). For temporally coherent results, our network works on space-time volumes of conditioning inputs. To process a complete video, we input the conditioning space-time volumes in a sliding window fashion, and assemble the final video from the output frames. We evaluate our approach (see Section 7) and show its potential on several video rewrite applications, such as full-head reenactment, gaze redirection, video dubbing, and interactive parameter-based video control.

## 4 MONOCULAR FACE RECONSTRUCTION

We employ a state-of-the-art dense face reconstruction approach that fits a parametric model of face and illumination to each video frame. It obtains a meaningful parametric face representation for the source $\mathcal{V}^s = \{\mathcal{I}_f^s \mid f = 1, \ldots, N_s\}$ and target $\mathcal{V}^t = \{\mathcal{I}_f^t \mid f = 1, \ldots, N_t\}$ video sequence, where $N_s$ and $N_t$ denote the total number of source and target frames, respectively. Let $\mathcal{P}^\bullet = \{\mathcal{P}_f^\bullet \mid f = 1, \ldots, N_\bullet\}$ be the corresponding parameter sequence that fully describes the source or target facial performance. The set of reconstructed parameters encode the rigid head pose (rotation $\mathbf{R}^\bullet \in SO(3)$ and translation $\mathbf{t}^\bullet \in \mathbb{R}^3$), facial identity coefficients $\boldsymbol{\alpha}^\bullet \in \mathbb{R}^{N_\alpha}$ (geometry, $N_\alpha = 80$) and $\boldsymbol{\beta}^\bullet \in \mathbb{R}^{N_\beta}$ (reflectance, $N_\beta = 80$), expression coefficients $\boldsymbol{\delta}^\bullet \in \mathbb{R}^{N_\delta}$ ($N_\delta = 64$), gaze direction for both eyes $\mathbf{e}^\bullet \in \mathbb{R}^4$, and spherical harmonics illumination coefficients $\boldsymbol{\gamma}^\bullet \in \mathbb{R}^{27}$. Overall, our monocular face tracker reconstructs $N_p = 261$ parameters per video frame. In the following, we provide more details on the face tracking algorithm as well as the parametric face representation.

*Parametric Face Representation.* We represent the space of facial identity based on a parametric head model [Blanz and Vetter 1999], and the space of facial expressions via an affine model. Mathematically, we model geometry variation through an affine model $\mathbf{v} \in \mathbb{R}^{3N}$ that stacks per-vertex deformations of the underlying template mesh with $N$ vertices, as follows:

$$\mathbf{v}(\boldsymbol{\alpha}, \boldsymbol{\delta}) = \mathbf{a}_{\text{geo}} + \sum_{k=1}^{N_\alpha} \alpha_k \mathbf{b}_k^{\text{geo}} + \sum_{k=1}^{N_\delta} \delta_k \mathbf{b}_k^{\text{exp}}. \quad (1)$$

Diffuse skin reflectance is modeled similarly by a second affine model $\mathbf{r} \in \mathbb{R}^{3N}$ that stacks the diffuse per-vertex albedo:

$$\mathbf{r}(\boldsymbol{\beta}) = \mathbf{a}_{\text{ref}} + \sum_{k=1}^{N_\beta} \beta_k \mathbf{b}_k^{\text{ref}}. \quad (2)$$

The vectors $\mathbf{a}_{\text{geo}} \in \mathbb{R}^{3N}$ and $\mathbf{a}_{\text{ref}} \in \mathbb{R}^{3N}$ store the average facial geometry and corresponding skin reflectance, respectively. The geometry basis $\{\mathbf{b}_k^{\text{geo}}\}_{k=1}^{N_\alpha}$ has been computed by applying principal component analysis (PCA) to 200 high-quality face scans [Blanz and Vetter 1999]. The reflectance basis $\{\mathbf{b}_k^{\text{ref}}\}_{k=1}^{N_\beta}$ has been obtained in the same manner. For dimensionality reduction, the expression basis $\{\mathbf{b}_k^{\text{exp}}\}_{k=1}^{N_\delta}$ has been computed using PCA, starting from the blendshapes of Alexander et al. [2010] and Cao et al. [2014b]. Their blendshapes have been transferred to the topology of Blanz and Vetter [1999] using deformation transfer [Sumner and Popović 2004].

*Image Formation Model.* To render synthetic head images, we assume a full perspective camera that maps model-space 3D points $\mathbf{v}$ via camera space $\hat{\mathbf{v}} \in \mathbb{R}^3$ to 2D points $\mathbf{p} = \Pi(\hat{\mathbf{v}}) \in \mathbb{R}^2$ on the image plane. The perspective mapping $\Pi$ contains the multiplication with the camera intrinsics and the perspective division. We assume a fixed and identical camera for all scenes, i.e., world and camera space are the same, and the face model accounts for all the scene motion. Based on a distant illumination assumption, we use the spherical harmonics (SH) basis functions $Y_b : \mathbb{R}^3 \to \mathbb{R}$ to approximate the incoming radiance $\mathbf{B}$ from the environment:

$$\mathbf{B}(\mathbf{r}_i, \mathbf{n}_i, \boldsymbol{\gamma}) = \mathbf{r}_i \cdot \sum_{b=1}^{B^2} \boldsymbol{\gamma}_b Y_b(\mathbf{n}_i). \quad (3)$$

Here, $B$ is the number of spherical harmonics bands, $\boldsymbol{\gamma}_b \in \mathbb{R}^3$ are the SH coefficients, and $\mathbf{r}_i$ and $\mathbf{n}_i$ are the reflectance and unit normal vector of the $i$-th vertex, respectively. For diffuse materials, an average approximation error below 1 percent is achieved with only $B = 3$ bands, independent of the illumination [Ramamoorthi and Hanrahan 2001], since the incident radiance is in general a smooth function. This results in $B^2 = 9$ parameters per color channel.

*Dense Face Reconstruction.* We employ a dense data-parallel face reconstruction approach to efficiently compute the parameters $\mathcal{P}^\bullet$ for both source and target videos. Face reconstruction is based on an *analysis-by-synthesis* approach that maximizes photo-consistency between a synthetic rendering of the model and the input. The reconstruction energy combines terms for dense photo-consistency, landmark alignment and statistical regularization:

$$E(\mathcal{X}) = w_{\text{photo}} E_{\text{photo}}(\mathcal{X}) + w_{\text{land}} E_{\text{land}}(\mathcal{X}) + w_{\text{reg}} E_{\text{reg}}(\mathcal{X}), \quad (4)$$

with $\mathcal{X} = \{\mathbf{R}^\bullet, \mathbf{t}^\bullet, \alpha^\bullet, \beta^\bullet, \delta^\bullet, \gamma^\bullet\}$. This enables the robust reconstruction of identity (geometry and skin reflectance), facial expression, and scene illumination. We use 66 automatically detected facial landmarks of the True Vision Solution tracker[1], which is a commercial implementation of Saragih et al. [2011], to define the sparse alignment term $E_\text{land}$. Similar to Thies et al. [2016], we use a robust $\ell_1$-norm for dense photometric alignment $E_\text{photo}$. The regularizer $E_\text{reg}$ enforces statistically plausible parameter values based on the assumption of normally distributed data. The eye gaze estimate $\mathbf{e}^\bullet$ is directly obtained from the landmark tracker. The identity is only estimated in the first frame and is kept constant afterwards. All other parameters are estimated every frame. For more details on the energy formulation, we refer to Garrido et al. [2016] and Thies et al. [2016]. We use a data-parallel implementation of iteratively re-weighted least squares (IRLS), similar to Thies et al. [2016], to find the optimal set of parameters. One difference to their work is that we compute and explicitly store the Jacobian $\mathbf{J}$ and the residual vector $\mathbf{F}$ to global memory based on a data-parallel strategy that launches one thread per matrix/vector element. Afterwards, a data-parallel matrix–matrix/matrix–vector multiplication computes the right- and left-hand side of the normal equations that have to be solved in each IRLS step. The resulting small linear system (97×97 in tracking mode, 6 DoF rigid pose, 64 expression parameters and 27 SH coefficients) is solved on the CPU using Cholesky factorization in each IRLS step. The reconstruction of a single frame takes 670 ms (all parameters) and 250 ms (without identity, tracking mode). This allows the efficient generation of the training corpus that is required by our space-time rendering-to-video translation network (see Section 6). Contrary to Garrido et al. [2016] and Thies et al. [2016], our model features dimensions to model eyelid closure, so eyelid motion is captured well.

## 5 SYNTHETIC CONDITIONING INPUT

Using the method from Section 4, we reconstruct the face in each frame of the source and unmodified target video. Next, we obtain the modified parameter vector for every frame of the target sequence, e.g., for full-head reenactment, we modify the rigid head pose, expression and eye gaze of the target actor. All parameters are copied in a relative manner from the source to the target, i.e., with respect to a neutral reference frame. Then we render synthetic conditioning images of the target actor's face model under the modified parameters using hardware rasterization. For higher temporal coherence, our rendering-to-video translation network takes a space-time volume of conditioning images $\{C_{f-o} \mid o = 0, \ldots, 10\}$ as input, with $f$ being the index of the current frame. We use a temporal window of size $N_w = 11$, with the current frame being at its end. This provides the network a history of the earlier motions.

For each frame $C_{f-o}$ of the window, we generate three different conditioning inputs: a color rendering, a correspondence image, and an eye gaze image (see Figure 3). The color rendering shows the modified target actor model under the estimated target illumination, while keeping the target identity (geometry and skin reflectance) fixed. This image provides a good starting point for the following rendering-to-video translation, since in the face region only the delta to a real image has to be learned. In addition to this color input, we also provide a correspondence image encoding the index of the parametric face model's vertex that projects into each pixel. To this end, we texture the head model with a constant unique gradient texture map, and render it. Finally, we also provide an eye gaze image that solely contains the white region of both eyes and the locations of the pupils as blue circles. This image provides information about the eye gaze direction and blinking to the network.

[1]http://truevisionsolutions.net

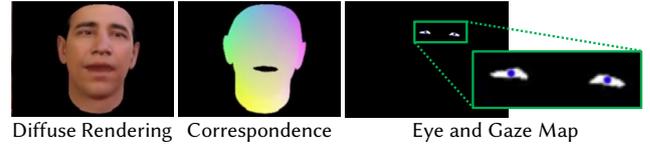

Fig. 3. The synthetic input used for conditioning our rendering-to-video translation network: (1) colored face rendering under target illumination, (2) correspondence image, and (3) the eye gaze image.

We stack all $N_w$ conditioning inputs of a time window in a 3D tensor $\mathbf{X}$ of size $W \times H \times 9N_w$ (3 images, with 3 channels each), to obtain the input to our rendering-to-video translation network. To process the complete video, we feed the conditioning space-time volumes in a sliding window fashion. The final generated photo-realistic video output is assembled directly from the output frames.

## 6 RENDERING-TO-VIDEO TRANSLATION

The generated conditioning space-time video tensors are the input to our rendering-to-video translation network. The network learns to convert the synthetic input into full frames of a photo-realistic target video, in which the target actor now mimics the head motion, facial expression and eye gaze of the synthetic input. The network learns to synthesize the entire actor in the foreground, i.e., the face for which conditioning input exists, but also all other parts of the actor, such as hair and body, so that they comply with the target head pose. It also synthesizes the appropriately modified and filled-in background, including even some consistent lighting effects between foreground and background. The network is trained for a specific target actor and a specific static, but otherwise general scene background. Our rendering-to-video translation network follows an encoder–decoder architecture and is trained in an adversarial manner based on a discriminator that is jointly trained. In the following, we explain the network architectures, the used loss functions and the training procedure in detail.

*Network Architecture.* We show the architecture of our rendering-to-video translation network in Figure 4. Our conditional generative adversarial network consists of a space-time transformation network $\mathbf{T}$ and a discriminator $\mathbf{D}$. The transformation network $\mathbf{T}$ takes the $W \times H \times 9N_w$ space-time tensor $\mathbf{X}$ as input and outputs a photo-real image $\mathbf{T}(\mathbf{X})$ of the target actor. The temporal input enables the network to take the history of motions into account by inspecting previous conditioning images. The temporal axis of the input tensor is aligned along the network channels, i.e., the convolutions in the first layer have $9N_w$ channels. Note, we store all image data in normalized $[-1, +1]$-space, i.e, black is mapped to $[-1, -1, -1]^\top$ and white is mapped to $[+1, +1, +1]^\top$.

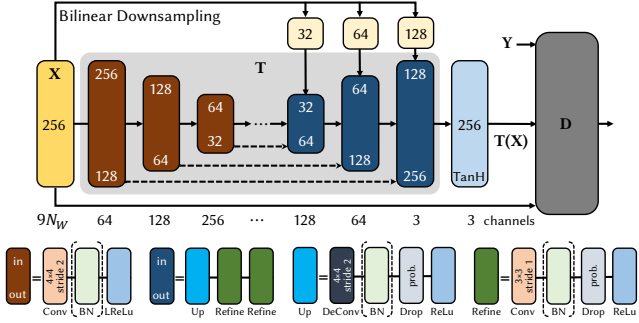

Fig. 4. Architecture of our rendering-to-video translation network for an input resolution of 256×256: The encoder has 8 downsampling modules with (64, 128, 256, 512, 512, 512, 512, 512) output channels. The decoder has 8 upsampling modules with (512, 512, 512, 512, 256, 128, 64, 3) output channels. The upsampling modules use the following dropout probabilities (0.5, 0.5, 0.5, 0, 0, 0, 0, 0). The first downsampling and the last upsampling module do not employ batch normalization (BN). The final non-linearity (TanH) brings the output to the employed normalized [−1, +1]-space.

Our network consists of two main parts, an encoder for computing a low-dimensional latent representation, and a decoder for synthesizing the output image. We employ skip connections [Ronneberger et al. 2015] to enable the network to transfer fine-scale structure. To generate video frames with sufficient resolution, our network also employs a cascaded refinement strategy [Chen and Koltun 2017]. In each downsampling step, we use a convolution (4 × 4, stride 2) followed by batch normalization and a leaky ReLU non-linearity. The upsampling module is specifically designed to produce high-quality output, and has the following structure: first, the resolution is increased by a factor of two based on deconvolution (4 × 4, upsampling factor of 2), batch normalization, dropout and ReLU. Afterwards, two refinement steps based on convolution (3 × 3, stride 1, stays on the same resolution) and ReLU are applied. The final hyperbolic tangent non-linearity (TanH) brings the output tensor to the normalized [−1, +1]-space used for storing the image data. For more details, please refer to Figure 4.

The input to our discriminator $D$ is the conditioning input tensor $X$ (size $W \times H \times 9N_w$), and either the predicted output image $T(X)$ or the ground-truth image, both of size $W \times H \times 3$. The employed discriminator is inspired by the PatchGAN classifier, proposed by Isola et al. [2017]. We extended it to take volumes of conditioning images as input.

*Objective Function.* We train in an adversarial manner to find the best rendering-to-video translation network:

$$T^* = \underset{T}{\text{argmin}} \max_{D} E_{\text{cGAN}}(T, D) + \lambda E_{\ell_1}(T). \quad (5)$$

This objective function comprises an adversarial loss $E_{\text{cGAN}}(T, D)$ and an $\ell_1$-norm reproduction loss $E_{\ell_1}(T)$. The constant weight of $\lambda = 100$ balances the contribution of these two terms. The adversarial loss has the following form:

$$E_{\text{GAN}}(T, D) = \mathbb{E}_{X,Y}\big[\log D(X, Y)\big] + \mathbb{E}_{X}\big[\log\big(1 - D(X, T(X))\big)\big]. \quad (6)$$

We do not inject a noise vector while training our network to produce deterministic outputs. During adversarial training, the discriminator $D$ tries to get better at classifying given images as *real* or *synthetic*, while the transformation network $T$ tries to improve in fooling the discriminator. The $\ell_1$-norm loss penalizes the distance between the synthesized image $T(X)$ and the ground-truth image $Y$, which encourages the sharpness of the synthesized output:

$$E_{\ell_1}(T) = \mathbb{E}_{X,Y}\big[\,\|Y - T(X)\|_1\,\big]. \quad (7)$$

*Training.* We construct the training corpus $\mathcal{T} = \{(X_i, Y_i)\}_i$ based on the tracked video frames of the target video sequence. Typically, two thousand video frames, i.e., about one minute of video footage, are sufficient to train our network (see Section 7). Our training corpus consists of $N_t - (N_w - 1)$ rendered conditioning space-time volumes $X_i$ and the corresponding ground-truth image $Y_i$ (using a window size of $N_w = 11$). We train our networks using the TensorFlow [Abadi et al. 2015] deep learning framework. The gradients for back-propagation are obtained using Adam [Kingma and Ba 2015]. We train for 31,000 iterations with a batch size of 16 (approx. 250 epochs for a training corpus of 2000 frames) using a base learning rate of 0.0002 and first momentum of 0.5; all other parameters have their default value. We train our networks from scratch, and initialize the weights based on a Normal distribution $\mathcal{N}(0, 0.2)$.

## 7 RESULTS

Our approach enables full-frame target video portrait synthesis under full 3D head pose control. We measured the runtime for training and testing on an Intel Xeon E5-2637 with 3.5 GHz (16 GB RAM) and an NVIDIA GeForce GTX Titan Xp (12 GB RAM). Training our network takes 10 hours for a target video resolution of 256×256 pixels, and 42 hours for 512×512 pixels. Tracking the source actor takes 250 ms per frame (without identity), and the rendering-to-video conversion (inference) takes 65 ms per frame for 256×256 pixels, or 196 ms for 512×512 pixels.

In the following, we evaluate the design choices of our deep video portrait algorithm, compare to current state-of-the-art reenactment approaches, and show the results of a large-scale web-based user study. We further demonstrate the potential of our approach on several video rewrite applications, such as reenactment under full head and facial expression control, facial expression reenactment only, video dubbing, and live video portrait editing under user control. In total, we applied our approach to 14 different target sequences of 13 different subjects and used 5 different source sequences; see Appendix A for details. A comparison to a simple nearest-neighbor retrieval approach can be found in Figure 6 and in the supplemental video. Our approach requires only a few minutes of target video footage for training.

### 7.1 Applications

Our approach enables us to take full control of the rigid head pose, facial expression, and eye motion of a target actor in a video portrait, thus opening up a wide range of video rewrite applications. All parameter dimensions can be estimated and transfered from a source video sequence or edited manually through an interactive user interface.

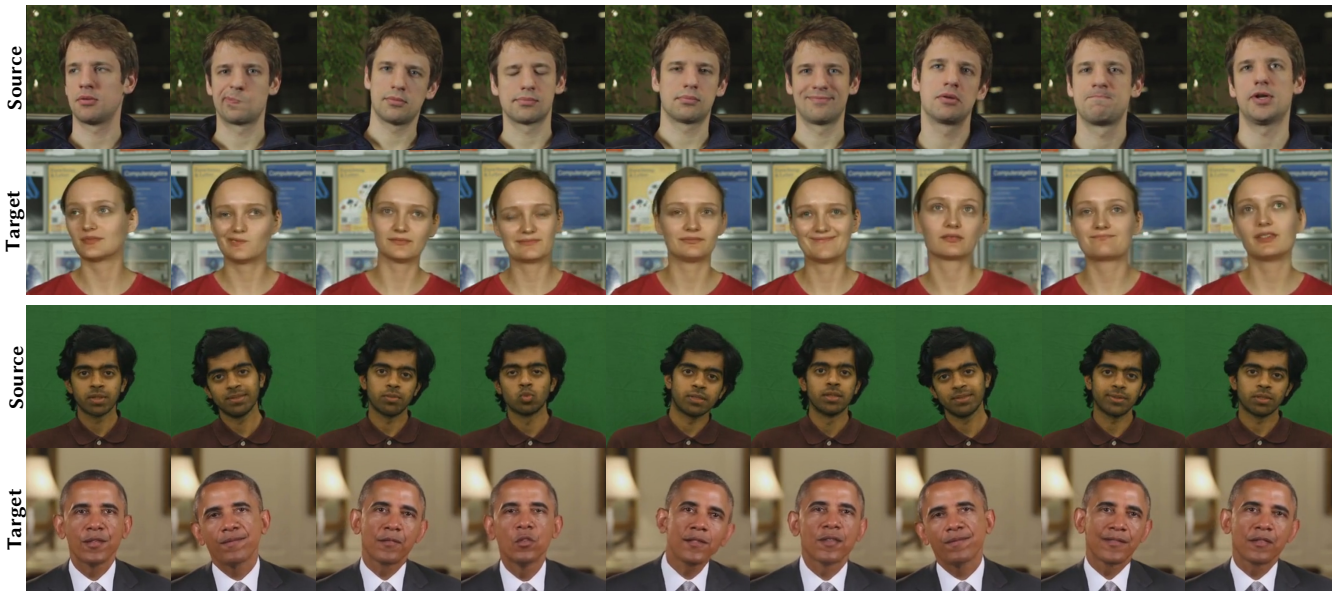

Fig. 5. Qualitative results of full-head reenactment: our approach enables full-frame target video portrait synthesis under full 3D head pose control. The output video portraits are photo-realistic and hard to distinguish from real videos. Note that even the shadow in the background of the second row moves consistently with the modified foreground head motion. In the sequence at the top, we only transfer the translation in the camera plane, while we transfer the full 3D translation for the sequence at the bottom. For full sequences, please refer to our video. *Obama* video courtesy of the White House (public domain).

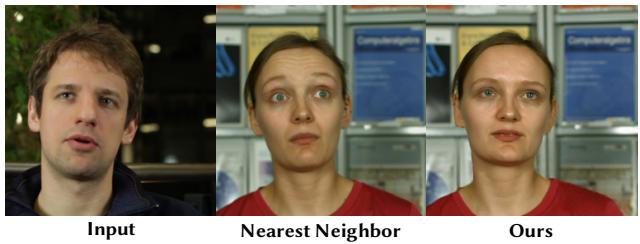

**Input**     **Nearest Neighbor**     **Ours**

Fig. 6. Comparison to a nearest-neighbor approach in parameter space (pose and expression). Our results have higher quality and are temporally more coherent (see supplemental video). For the nearest-neighbor approach, it is difficult to find the right trade-off between pose and expression. This leads to many results with one of the two dimensions not being well-matched. The results are also temporally unstable, since the nearest neighbor abruptly changes, especially for small training sets.

*Reenactment under full head control.* Our approach is the first that can photo-realistically transfer the full 3D head pose (spatial position and rotation), facial expression, as well as eye gaze and eye blinking of a captured source actor to a target actor video. Figure 5 shows some examples of full-head reenactment between different source and target actors. Here, we use the full target video for training and the source video as the driving sequence. As can be seen, the output of our approach achieves a high level of realism and faithfully mimics the driving sequence, while still retaining the mannerisms of the original target actor. Note that the shadow in the background moves consistently with the position of the actor in the scene, as shown in Figure 5 (second row). We also demonstrate the high quality of our results and evaluate our approach quantitatively in a self-reenactment scenario, see Figure 7. For the quantitative analysis, we use two thirds of the target video for training and one third for testing. We capture the face in the training and driving video with our model-based tracker, and then render the conditioning images, which serve as input to our network for synthesizing the output. For further details, please refer to Section 7.2. Note that the synthesized results are nearly indistinguishable from the ground truth.

*Facial Reenactment and Video Dubbing.* Besides full-head reenactment, our approach also enables facial reenactment. In this experiment, we replace the expression coefficients of the target actor with those of the source actor before synthesizing the conditioning input to our rendering-to-video translation network. Here, the head pose and position, and eye gaze remain unchanged. Figure 8 shows facial reenactment results. Observe that the face expression in the synthesized target video nicely matches the expression of the source actor in the driving sequence. Please refer to the supplemental video for the complete video sequences.

Our approach can also be applied to visual dubbing. In many countries, foreign-language movies are dubbed, i.e., the original voice of an actor is replaced with that of a dubbing actor speaking in another language. Dubbing often causes visual discomfort due to the discrepancy between the actor's mouth motion and the new audio track. Even professional dubbing studios achieve only approximate audio alignment at best. Visual dubbing aims at altering the mouth motion of the target actor to match the new foreign-language audio track spoken by the dubber. Figure 9 shows results where we modify the facial motion of actors speaking originally in German to adhere to an English translation spoken by a professional dubbing actor, who was filmed in a dubbing studio [Garrido et al. 2015]. More precisely, we transfer the captured facial expressions of the dubbing actor to the target actor, while leaving the original target gaze and eye blinks intact, i.e., we use the original eye gaze images of the tracked target sequence as conditioning. As can be seen, our approach achieves dubbing results of high quality. In fact, we produce images with more realistic mouth interior and more emotional content in the mouth region. Please see the supplemental video for full video results.

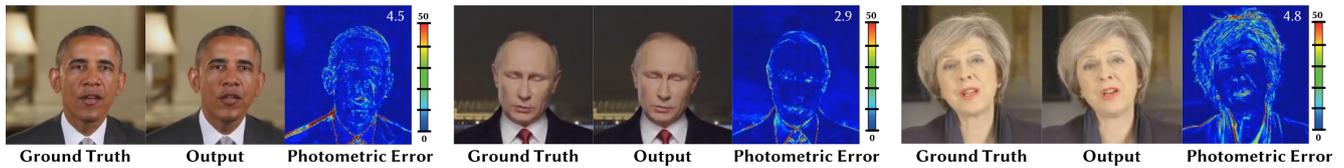

Fig. 7. Quantitative evaluation of the photometric re-rendering error. We evaluate our approach quantitatively in a self-reenactment setting, where the ground-truth video portrait is known. We train our rendering-to-video translation network on two thirds of the video sequence, and test on the remaining third. The error maps show per-pixel Euclidean distance in RGB (color channels in [0, 255]); the mean photometric error of the test set is shown in the top-right. The error is consistently low in regions with conditioning input, with higher errors in regions without conditioning, such as the upper body. *Obama* video courtesy of the White House (public domain). *Putin* video courtesy of the Kremlin (CC BY). *May* video courtesy of the UK government (Open Government Licence).

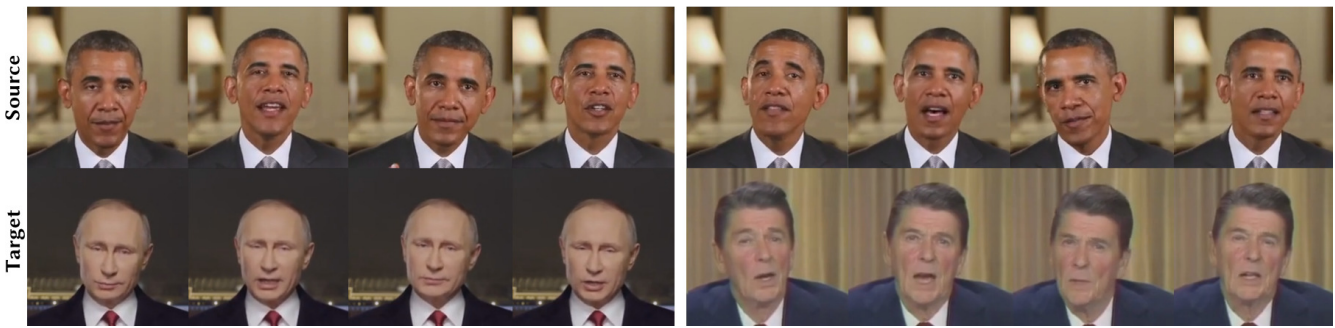

Fig. 8. Facial reenactment results of our approach. We transfer the expressions from the source to the target actor, while retaining the head pose (rotation and translation) as well as the eye gaze of the target actor. For the full sequences, please refer to the supplemental video. *Obama* video courtesy of the White House (public domain). *Putin* video courtesy of the Kremlin (CC BY). *Reagan* video courtesy of the National Archives and Records Administration (public domain).

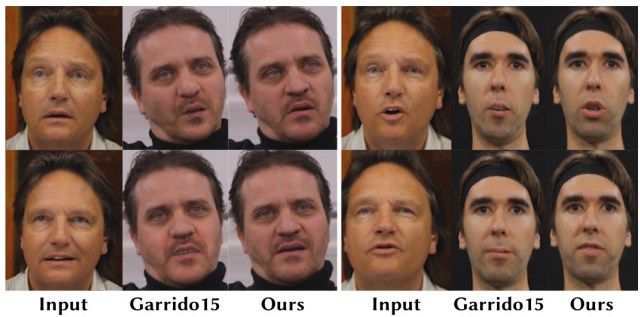

Fig. 9. Dubbing comparison on two sequences of Garrido et al. [2015]. For visual dubbing, we transfer the facial expressions of the dubbing actor ('input') to the target actor. We compare our results to Garrido et al.'s. Our approach obtains higher quality results in terms of the synthesized mouth shape and mouth interior. Note that our approach also enables full-head reenactment in addition to expression transfer. For the full comparison, we refer to the supplemental video.

*Interactive Editing of Video Portraits.* We built an interactive editor that enables users to reanimate video portraits with live feedback by modifying the parameters of the coarse face model rendered into the conditioning images (see our live demo in the supplemental video). Figure 10 shows a few static snapshots that were taken while the users were playing with our editor. Our approach enables changes of all parameter dimensions, either independently or all together, as shown in Figure 10. More specifically, we show independent changes of the expression, head rotation, head translation, and eye gaze (including eye blinks). Please note the realistic and consistent generation of the torso, head and background. Even shadows or reflections appear very consistently in the background. In addition, we show user edits that modify all parameters simultaneously. Our interactive editor runs at approximately 9 fps. While not the focus of this paper, our approach also enables modifications of the geometric facial identity, see Figure 11. These combined modifications show as a proof of concept that our network generalizes beyond the training corpus.

### 7.2 Quantitative Evaluation

We performed a quantitative evaluation of the re-rendering quality. First, we evaluate our approach in a self-reenactment setting, where the ground-truth video portrait is known. We train our rendering-to-video translation network on the first two thirds of a video sequence and test it on the remaining last third of the video, see Figure 7. The photometric error maps show the per-pixel Euclidean distance in RGB color space, with each channel being in [0, 255]. We performed this test for three different videos and the mean photometric errors are 2.88 (Vladimir Putin), 4.76 (Theresa May), and 4.46 (Barack Obama). Our approach obtains consistently low error in regions with conditioning input (face) and higher errors are found in regions that are unexplained by the conditioning input. Please note that while the synthesized video portraits slightly differ from the ground truth outside the face region, the synthesized hair and upper body are still plausible, consistent with the face region, and free of visual artifacts. For a complete analysis of these sequences, we refer to the supplemental video.

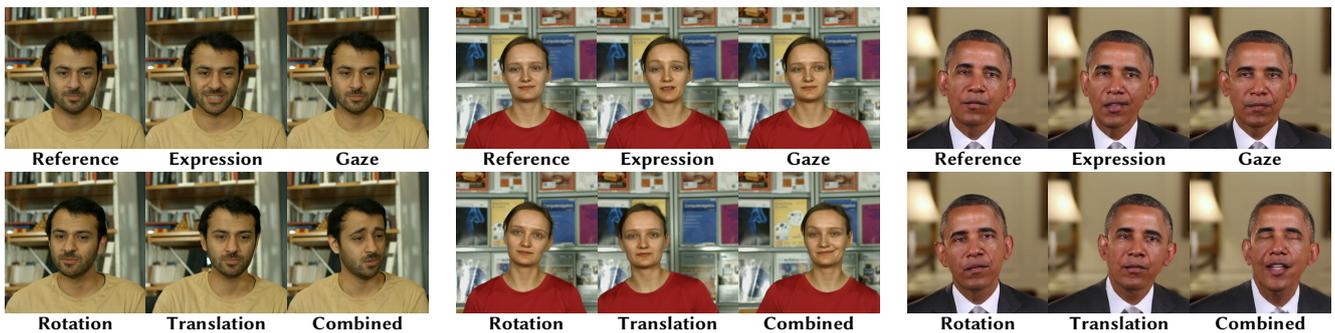

Fig. 10. Interactive editing. Our approach provides full parametric control over video portraits (by controlling head model parameters in conditioning images). This enables modifications of the rigid head pose (rotation and translation), facial expression and eye motion. All of these dimensions can be manipulated together or independently. We also show these modifications live in the supplemental video. *Obama* video courtesy of the White House (public domain).

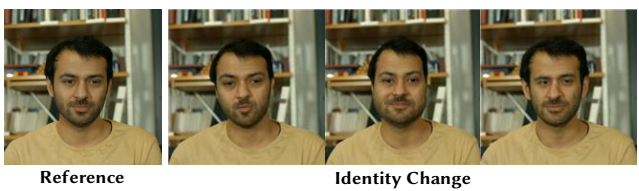

Fig. 11. Identity modification. While not the main focus of our approach, it also enables modification of the facial shape via the geometry shape parameters. This shows that our network picks up the correspondence between the model and the video portrait. Note that the produced outputs are also consistent in regions that are not constrained by the conditioning input, such as the hair and background.

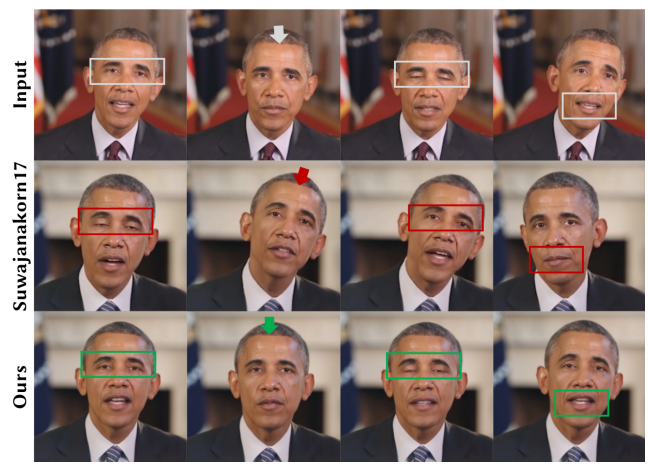

Fig. 13. Comparison to Suwajanakorn et al. [2017]. Their approach produces accurate lip sync with visually imperceptible artifacts, but provides no direct control over facial expressions. Thus, the expressions in the output do not always perfectly match the input (box, mouth), especially for expression changes without audio cue. Our visual dubbing approach accurately transfers the expressions from the source to the target. In addition, our approach provides more control over the target video by also transferring the eye gaze and eye blinks (box, eyes), and the rigid head pose (arrows). Since the source sequence shows more head-pose variation than the target sequence, we scaled the transferred rotation and translation by 0.5 in this experiment. For the full video sequence, we refer to the supplemental video. *Obama* video courtesy of the White House (public domain).

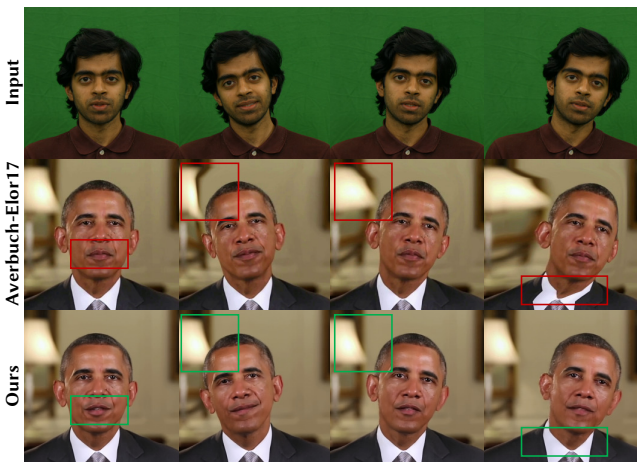

Fig. 12. Comparison to the image reenactment approach of Averbuch-Elor et al. [2017] in the full-head reenactment scenario. Since their method is based on a single target image, they copy the mouth interior from the source to the target, thus not preserving the target's identity. Our learning-based approach enables larger modifications of the rigid head pose without apparent artifacts, while their warping-based approach distorts the head and background. In addition, ours enables joint control of the eye gaze and eye blinks. The differences are most evident in the supplemental video. *Obama* video courtesy of the White House (public domain).

We evaluate our space-time conditioning strategy in Figure 16. Without space-time conditioning, the photometric error is significantly higher. The average errors over the complete sequence are 4.9 without vs. 4.5 with temporal conditioning (Barack Obama) and 5.3 without vs. 4.8 with temporal conditioning (Theresa May). In addition to a lower photometric error, space-time conditioning also leads to temporally significantly more stable video outputs. This can be seen best in the supplemental video.

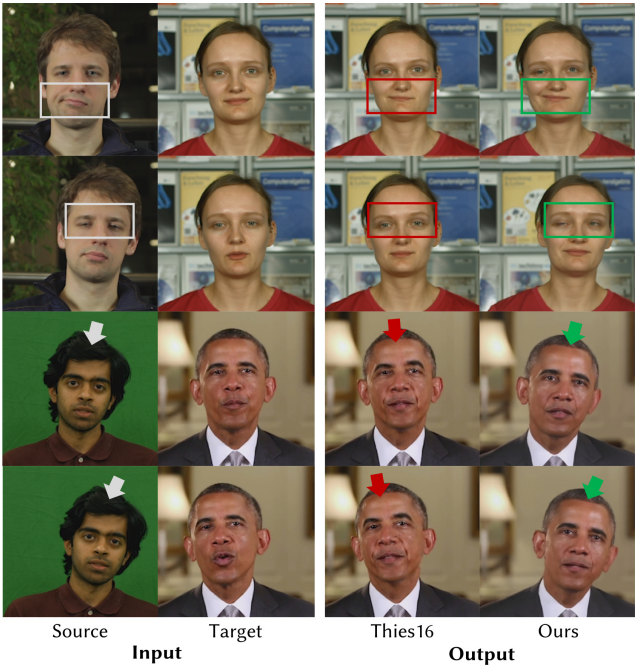

Fig. 14. Comparison to the state-of-the-art facial reenactment approach of Thies et al. [2016]. Our approach achieves expression transfer of similar quality, while also enabling full-head reenactment, i.e., it also transfers the rigid head pose, gaze direction, and eye blinks. For the video result, we refer to the supplemental video. *Obama* video courtesy of the White House (public domain).

We also evaluate the importance of the training set size. In this experiment, we train our rendering-to-video translation network with 500, 1000, 2000 and 4000 frames of the target sequence, see Figure 15. As can be expected, larger training sets produce better results, and the best results are obtained with the full training set.

We also evaluate different image resolutions by training our rendering-to-video translation network for resolutions of 256×256, 512×512 and 1024×1024 pixels. We evaluate the quality in the self-reenactment setting, as shown in Figure 17. Generative networks of higher resolution are harder to train and require significantly longer training times: 10 hours for 256×256, 42 hours for 512×512, and 110 hours for 1024×1024 (on a Titan Xp). Therefore, we use a resolution of 256×256 pixels for most results.

## 7.3 Comparisons to the State of the Art

We compare our deep video portrait approach to current state-of-the-art video and image reenactment techniques.

*Comparison to Thies et al. [2016].* We compare our approach to the state-of-the-art *Face2Face* facial reenactment method of Thies et al. [2016]. In comparison to *Face2Face*, our approach achieves expression transfer of similar quality. What distinguishes our approach is the capability for full-head reenactment, i.e., the ability to also transfer the rigid head pose, gaze direction, and eye blinks in addition to the facial expressions, as shown in Figure 14. As can be seen, in our result, the head pose and eye motion nicely matches the source sequence, while the output generated by *Face2Face* follows the head and eye motion of the original target sequence. Please see the supplemental video for the video result.

*Comparison to Suwajanakorn et al. [2017].* We also compare to the audio-based dubbing approach of Suwajanakorn et al. [2017], see Figure 13. Their *AudioToObama* approach produces accurate lip sync with visually imperceptible artifacts, but provides no direct control over facial expressions. Thus, the expressions in the output do not always perfectly match the input (box, mouth), especially for expression changes without an audio cue. Our visual dubbing approach accurately transfers the expressions from the source to the target. In addition, our approach provides more control over the target video by also transferring the eye gaze and eye blinks (box, eyes) and the general rigid head pose (arrows). While their approach is trained on a huge amount of training data (17 hours), our approach only uses a small training dataset (1.3 minutes). The differences are best visible in the supplemental video.

*Comparison to Averbuch-Elor et al. [2017].* We compare our approach in the full-head reenactment scenario to the image reenactment approach of Averbuch-Elor et al. [2017], see Figure 12. Their approach does not preserve the identity of the target actor, since they copy the teeth and mouth interior from the source to the target sequence. Our learning-based approach enables larger modifications of the head pose without apparent artifacts, while their warping-based approach significantly distorts the head and background. In addition, we enable the joint modification of the gaze direction and eye blinks; see supplemental video.

## 7.4 User Study

We conducted two extensive web-based user studies to quantitatively evaluate the realism of our results. We prepared short 5-second video clips that we extracted from both real and synthesized videos (see Figure 18), to evaluate three applications of our approach: self-reenactment, same-person-reenactment and visual dubbing. We opted for self-reenactment, same-person-reenactment (two speeches of Barack Obama) and visual dubbing to guarantee that the motion types in the evaluated real and synthesized video pairs are matching. This eliminates the motion type as a confounding factor from the statistical analysis, e.g., having unrealistic motions for a public speech in the synthesized videos would negatively bias the outcome of the study. Our evaluation is focused on the visual quality of the synthesized results. Most video clips have a resolution of 256×256 pixels, but some are 512×512 pixels. In our user study, we presented one video clip at a time, and asked participants to respond to the statement "*This video clip looks real to me*" on a 5-point Likert scale (1–*strongly disagree*, 2–*disagree*, 3–*don't know*, 4–*agree*, 5–*strongly agree*). Video clips are shown in a random order, and each video clip is shown exactly once to assess participants' first impression. We recruited 135 and 69 anonymous participants for our two studies, largely from North America and Europe.

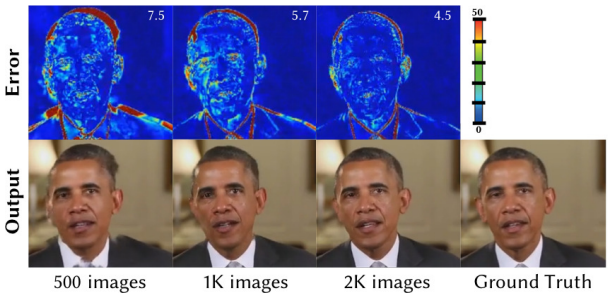
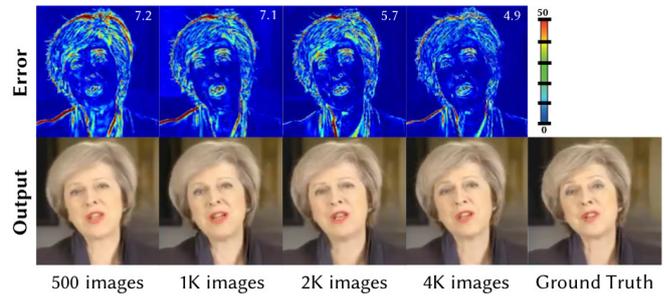

Fig. 15. Quantitative evaluation of the training set size. We train our rendering-to-video translation network with training corpora of different sizes. The error maps show the per-pixel distance in RGB color space with each channel being in [0, 255]; the mean photometric error is shown in the top-right. Smaller training sets have larger photometric errors, especially for regions outside of the face. For the full comparison, we refer to the supplemental video. *Obama* video courtesy of the White House (public domain). *May* video courtesy of the UK government (Open Government Licence).

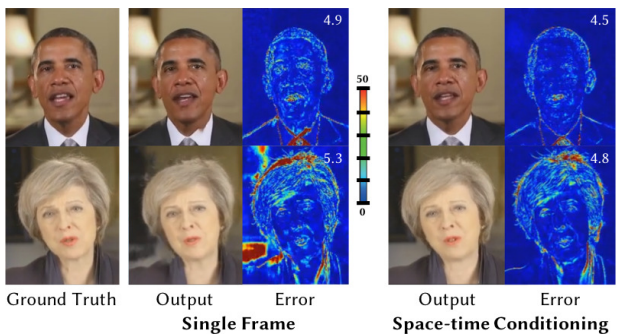

Fig. 16. Quantitative evaluation of the influence of the proposed space-time conditioning input. The error maps show the per-pixel distance in RGB color space with each channel being in [0, 255]; the mean photometric error is shown in the top-right. Without space-time conditioning, the photometric error is higher. Temporal conditioning adds significant temporal stability. This is best seen in the supplemental video. *Obama* video courtesy of the White House (public domain). *May* video courtesy of the UK government (Open Government Licence).

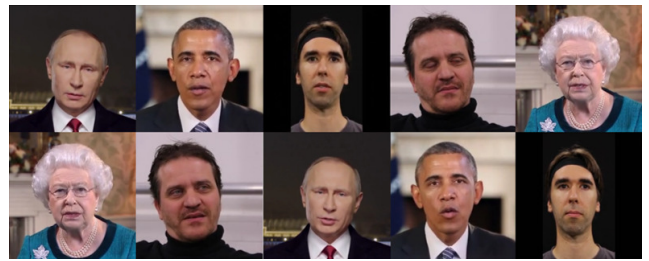

Fig. 18. We performed a user study to evaluate the quality of our results and see if users can distinguish between real (top) and synthesized video clips (bottom). The video clips include self-reenactment, same-person-reenactment, and video dubbing. *Putin* video courtesy of the Kremlin (CC BY). *Obama* video courtesy of the White House (public domain). *Elizabeth II* video courtesy of the Governor General of Canada (public domain).

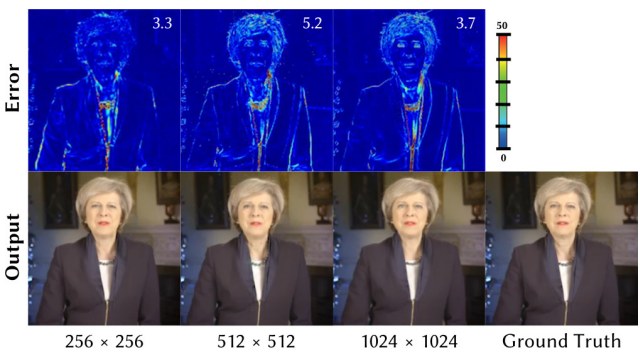

Fig. 17. Quantitative comparison of different resolutions. We train three rendering-to-video translation networks for resolutions of 256×256, 512×512 and 1024×1024 pixels. The error maps show the per-pixel distance in RGB color space with each channel being in [0, 255]; the mean photometric error is shown in the top-right. For the full comparison, see our video. *May* video courtesy of the UK government (Open Government Licence).

The results in Table 1 show that only 80% of participants rated real 256×256 videos as real, i.e. (strongly) agreeing to the video looking real; it seems that in anticipation of synthetic video clips, participants became overly critical. At the same time, 50% of participants consider our 256×256 results to be real, which increases slightly to 52% for 512×512. Our best result is the self-reenactment of Vladimir Putin at 256×256 resolution, which 65% of participants consider to be real, compared to 78% for the real video. We also evaluated partial and full reenactment by transferring a speech by Barack Obama to another video clip of himself. Table 2 indicates that we achieve better realism ratings with full reenactment comprising facial expressions and pose (50%) compared to transferring only facial expressions (38%). This might be because full-head reenactment keeps expressions and head motion synchronized. Suwajanakorn et al.'s speech-driven reenactment approach [2017] achieves a realism rating of 64% compared to the real source and target video clips, which achieve 70–86%. Our full-head reenactment results are considered to be at least as real as Suwajanakorn et al.'s by 60% of participants. We finally compared our dubbing results to VDub [Garrido et al. 2015] in Table 3. Overall, 57% of participants gave our results a higher realism rating (and 32% gave the same rating). Our results are again considered to be real by 51% of participants, compared to only 21% for VDub.

Table 1. User study results for self-reenacted videos ($n=135$). Columns 1–5 show the percentage of ratings given about the statement "This video clip looks real to me", from 1 (*strongly disagree*) to 5 (*strongly agree*). 4+5='real'.

|  | res | Real videos |  |  |  |  |  | Our results |  |  |  |  |  |
| --- | --- | --- | --- | --- | --- | --- | --- | --- | --- | --- | --- | --- | --- |
|  |  | 1 | 2 | 3 | 4 | 5 | 'real' | 1 | 2 | 3 | 4 | 5 | 'real' |
| Obama | 256 | 2 | 8 | 10 | 62 | 19 | 81% | 13 | 33 | 11 | 37 | 6 | 43% |
| Putin | 256 | 2 | 11 | 10 | 58 | 20 | 78% | 3 | 17 | 15 | 54 | 11 | 65% |
| Eliabeth II | 256 | 2 | 6 | 12 | 59 | 21 | 80% | 6 | 32 | 20 | 33 | 9 | 42% |
| Obama | 512 | 0 | 7 | 3 | 49 | 42 | 91% | 9 | 35 | 13 | 36 | 8 | 44% |
| Putin | 512 | 4 | 13 | 10 | 47 | 25 | 72% | 2 | 20 | 15 | 44 | 19 | 63% |
| Eliabeth II | 512 | 1 | 7 | 4 | 55 | 34 | 89% | 7 | 33 | 10 | 38 | 13 | 51% |
| Mean | 256 | 2 | 8 | 10 | 60 | 20 | 80% | 7 | 27 | 15 | 41 | 9 | 50% |
| Mean | 512 | 2 | 9 | 6 | 50 | 34 | 84% | 6 | 29 | 12 | 39 | 13 | 52% |

Table 2. User study results for expression and full head transfer between two videos of Barack Obama compared to the input videos and Suwajanakorn et al.'s approach ($n=69$, mean of 4 clips).

|  | Ratings |  |  |  |  |  |
| --- | --- | --- | --- | --- | --- | --- |
|  | 1 | 2 | 3 | 4 | 5 | 'real' |
| Source video (real) | 0 | 8 | 6 | 43 | 42 | 86% |
| Target video (real) | 1 | 14 | 14 | 47 | 23 | 70% |
| Suwajanakorn et al. [2017] | 2 | 20 | 14 | 47 | 17 | 64% |
| Expression transfer (ours) | 9 | 37 | 17 | 29 | 9 | 38% |
| Full head transfer (ours) | 3 | 31 | 16 | 37 | 13 | 50% |

Table 3. User study results for dubbing comparison to VDub ($n=135$).

|  | Garrido et al. [2015] |  |  |  |  |  | Our results |  |  |  |  |  |
| --- | --- | --- | --- | --- | --- | --- | --- | --- | --- | --- | --- | --- |
|  | 1 | 2 | 3 | 4 | 5 | 'real' | 1 | 2 | 3 | 4 | 5 | 'real' |
| Ingmar (3 clips) | 21 | 36 | 21 | 20 | 2 | 22% | 4 | 21 | 25 | 42 | 8 | 50% |
| Thomas (3 clips) | 33 | 36 | 11 | 16 | 4 | 20% | 7 | 25 | 17 | 42 | 9 | 51% |
| Mean (6 clips) | 27 | 36 | 16 | 18 | 3 | 21% | 6 | 23 | 21 | 42 | 9 | 51% |

On average, across all scenarios and both studies, our results are considered to be real by 47% of the participants (1,767 ratings), compared to only 80% for real video clips (1,362 ratings). This suggests that our results already fool about 60% of the participants – a good result given the critical participant pool. However, there is some variation across our results: lower realism ratings were given for well-known personalities like Barack Obama, while higher ratings were given for instance to the unknown dubbing actors.

## 8 DISCUSSION

While we have demonstrated highly realistic reenactment results in a large variety of applications and scenarios, our approach is also subject to a few limitations. Similar to all other learning-based approaches, ours works very well inside the span of the training corpus. Extreme target head poses, such as large rotations, or expressions far outside this span can lead to a degradation of the visual quality of the generated video portrait, see Figure 19 and the supplemental video. Since we only track the face with a parametric model, we cannot actively control the motion of the torso or hair, or control the background. The network learns to extrapolate and finds a plausible and consistent upper body and background (including some shadows and reflections) for a given head pose. This limitation could be overcome by also tracking the body and using the underlying body model to generate an extended set of conditioning images. Currently, we are only able to produce medium-resolution output due to memory and training time limitations. The limited output resolution makes it especially difficult to reproduce fine-scale detail, such as individual teeth, in a temporally coherent manner. Yet, recent progress on high-resolution discriminative adversarial networks [Karras et al. 2018; Wang et al. 2017] is promising and could be leveraged to further increase the resolution of the generated output. On a broader scale, and not being a limitation, democratization of advanced high-quality video editing possibilities, offered by our and other methods, calls for additional care in ensuring verifiable video authenticity, e.g., through invisible watermarking.

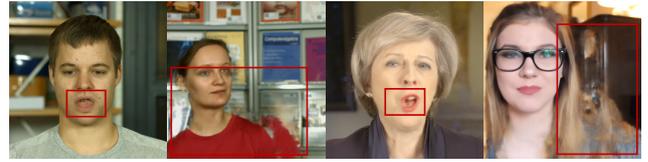

Fig. 19. Our approach works well within the span of the training corpus. Extreme changes in head pose far outside the training set or strong changes to the facial expression might lead to artifacts in the synthesized video. This is a common limitation of all learning-based approaches. In these cases, artifacts are most prominent outside the face region, as these regions have no conditioning input. *May* video courtesy of the UK government (Open Government Licence). *Malou* video courtesy of Louisa Malou (CC BY).

## 9 CONCLUSION

We presented a new approach to synthesize entire photo-real video portraits of a target actors in front of general static backgrounds. It is the first to transfer head pose and orientation, face expression, and eye gaze from a source actor to a target actor. The proposed method is based on a novel rendering-to-video translation network that converts a sequence of simple computer graphics renderings into photo-realistic and temporally-coherent video. This mapping is learned based on a novel space-time conditioning volume formulation. We have shown through experiments and a user study that our method outperforms prior work in quality and expands over their possibilities. It thus opens up a new level of capabilities in many applications, like video reenactment for virtual reality and telepresence, interactive video editing, and visual dubbing. We see our approach as a step towards highly realistic synthesis of full-frame video content under control of meaningful parameters. We hope that it will inspire future research in this very challenging field.

## ACKNOWLEDGMENTS

We are grateful to all our actors. We thank True-VisionSolutions Pty Ltd for kindly providing the 2D face tracker and Adobe for a Premiere Pro CC license. We also thank Supasorn Suwajanakorn and Hadar Averbuch-Elor for the comparisons. This work was supported by ERC Starting Grant CapReal (335545), a TUM-IAS Rudolf Mößbauer Fellowship, a Google Faculty Award, RCUK grant CAMERA (EP/M023281/1), an NVIDIA Corporation GPU Grant, and the Max Planck Center for Visual Computing and Communications (MPC-VCC).

# A APPENDIX

This appendix describes all the used datasets, see Table 4 (target actors) and Table 5 (source actors).

Table 4. Target videos: Name and length of sequences (in frames). *Malou* video courtesy of Louisa Malou (CC BY). *May* video courtesy of the UK government (Open Government Licence). *Obama* video courtesy of the White House (public domain). *Putin* video courtesy of the Kremlin (CC BY). *Reagan* video courtesy of the National Archives and Records Administration (public domain). *Elizabeth II* video courtesy of the Governor General of Canada (public domain). *Reagan* video courtesy of the National Archives and Records Administration (public domain). *Wolf* video courtesy of Tom Wolf (CC BY).

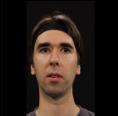
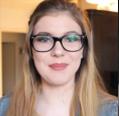
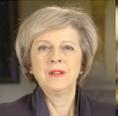
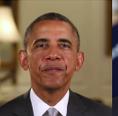
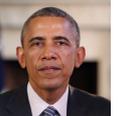

| Ingmar | Malou | May | Obama1 | Obama2 |
| --- | --- | --- | --- | --- |
| 3,000 | 15,000 | 5,000 | 2,000 | 3,613 |

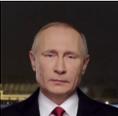
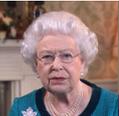
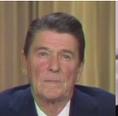
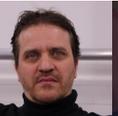
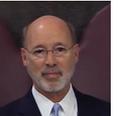

| Putin | Elizabeth II | Reagan | Thomas | Wolf |
| --- | --- | --- | --- | --- |
| 4,000 | 1,500 | 6,984 | 2,239 | 15,000 |

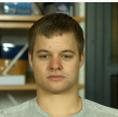
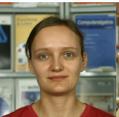
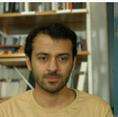
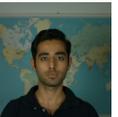

| DB1 | DB2 | DB3 | DB4 |
| --- | --- | --- | --- |
| 8,000 | 18,138 | 6,500 | 30,024 |

Table 5. Source videos: Name and length of sequences (in frames). *Obama* video courtesy of the White House (public domain).

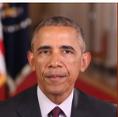
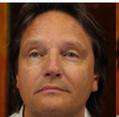
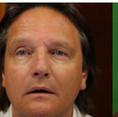
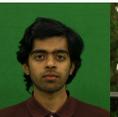
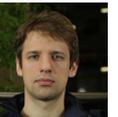

| Obama3 | David1 | David2 | DB5 | DB6 |
| --- | --- | --- | --- | --- |
| 1,945 | 4,611 | 3,323 | 3,824 | 2,380 |

# REFERENCES

Martín Abadi, Ashish Agarwal, Paul Barham, Eugene Brevdo, Zhifeng Chen, Craig Citro, Greg S. Corrado, Andy Davis, Jeffrey Dean, Matthieu Devin, Sanjay Ghemawat, Ian Goodfellow, Andrew Harp, Geoffrey Irving, Michael Isard, Yangqing Jia, Rafal Jozefowicz, Lukasz Kaiser, Manjunath Kudlur, Josh Levenberg, Dan Mané, Rajat Monga, Sherry Moore, Derek Murray, Chris Olah, Mike Schuster, Jonathon Shlens, Benoit Steiner, Ilya Sutskever, Kunal Talwar, Paul Tucker, Vincent Vanhoucke, Vijay Vasudevan, Fernanda Viégas, Oriol Vinyals, Pete Warden, Martin Wattenberg, Martin Wicke, Yuan Yu, and Xiaoqiang Zheng. 2015. TensorFlow: Large-Scale Machine Learning on Heterogeneous Systems. https://www.tensorflow.org/ Software available from tensorflow.org.

Oleg Alexander, Mike Rogers, William Lambeth, Jen-Yuan Chiang, Wan-Chun Ma, Chuan-Chang Wang, and Paul Debevec. 2010. The Digital Emily Project: Achieving a Photorealistic Digital Actor. *IEEE Computer Graphics and Applications* 30, 4 (July/August 2010), 20–31. https://doi.org/10.1109/MCG.2010.65

Hadar Averbuch-Elor, Daniel Cohen-Or, Johannes Kopf, and Michael F. Cohen. 2017. Bringing Portraits to Life. *ACM Transactions on Graphics (SIGGRAPH Asia)* 36, 6 (November 2017), 196:1–13. https://doi.org/10.1145/3130800.3130818

Volker Blanz, Kristina Scherbaum, Thomas Vetter, and Hans-Peter Seidel. 2004. Exchanging Faces in Images. *Computer Graphics Forum (Eurographics)* 23, 3 (September 2004), 669–676. https://doi.org/10.1111/j.1467-8659.2004.00799.x

Volker Blanz and Thomas Vetter. 1999. A Morphable Model for the Synthesis of 3D Faces. In *Annual Conference on Computer Graphics and Interactive Techniques (SIGGRAPH)*. 187–194. https://doi.org/10.1145/311535.311556

James Booth, Anastasios Roussos, Allan Ponniah, David Dunaway, and Stefanos Zafeiriou. 2018. Large Scale 3D Morphable Models. *International Journal of Computer Vision* 126, 2 (April 2018), 233–254. https://doi.org/10.1007/s11263-017-1009-7

Christoph Bregler, Michele Covell, and Malcolm Slaney. 1997. Video Rewrite: Driving Visual Speech with Audio. In *Annual Conference on Computer Graphics and Interactive Techniques (SIGGRAPH)*. 353–360. https://doi.org/10.1145/258734.258880

Chen Cao, Derek Bradley, Kun Zhou, and Thabo Beeler. 2015. Real-time High-fidelity Facial Performance Capture. *ACM Transactions on Graphics (SIGGRAPH)* 34, 4 (July 2015), 46:1–9. https://doi.org/10.1145/2766943

Chen Cao, Qiming Hou, and Kun Zhou. 2014a. Displaced Dynamic Expression Regression for Real-time Facial Tracking and Animation. *ACM Transactions on Graphics (SIGGRAPH)* 33, 4 (July 2014), 43:1–10. https://doi.org/10.1145/2601097.2601204

Chen Cao, Yanlin Weng, Shun Zhou, Yiying Tong, and Kun Zhou. 2014b. FaceWarehouse: A 3D Facial Expression Database for Visual Computing. *IEEE Transactions on Visualization and Computer Graphics* 20, 3 (March 2014), 413–425. https://doi.org/10.1109/TVCG.2013.249

Chen Cao, Hongzhi Wu, Yanlin Weng, Tianjia Shao, and Kun Zhou. 2016. Real-time Facial Animation with Image-based Dynamic Avatars. *ACM Transactions on Graphics (SIGGRAPH)* 35, 4 (July 2016), 126:1–12. https://doi.org/10.1145/2897824.2925873

Yao-Jen Chang and Tony Ezzat. 2005. Transferable Videorealistic Speech Animation. In *Symposium on Computer Animation (SCA)*. 143–151. https://doi.org/10.1145/1073368.1073388

Qifeng Chen and Vladlen Koltun. 2017. Photographic Image Synthesis with Cascaded Refinement Networks. In *International Conference on Computer Vision (ICCV)*. 1520–1529. https://doi.org/10.1109/ICCV.2017.168

Kevin Dale, Kalyan Sunkavalli, Micah K. Johnson, Daniel Vlasic, Wojciech Matusik, and Hanspeter Pfister. 2011. Video face replacement. *ACM Transactions on Graphics (SIGGRAPH Asia)* 30, 6 (December 2011), 130:1–10. https://doi.org/10.1145/2070781.2024164

Tony Ezzat, Gadi Geiger, and Tomaso Poggio. 2002. Trainable Videorealistic Speech Animation. *ACM Transactions on Graphics (SIGGRAPH)* 21, 3 (July 2002), 388–398. https://doi.org/10.1145/566654.566594

Ohad Fried, Eli Shechtman, Dan B. Goldman, and Adam Finkelstein. 2016. Perspective-aware Manipulation of Portrait Photos. *ACM Transactions on Graphics (SIGGRAPH)* 35, 4 (July 2016), 128:1–10. https://doi.org/10.1145/2897824.2925933

Graham Fyffe, Andrew Jones, Oleg Alexander, Ryosuke Ichikari, and Paul Debevec. 2014. Driving High-Resolution Facial Scans with Video Performance Capture. *ACM Transactions on Graphics* 34, 1 (December 2014), 8:1–14. https://doi.org/10.1145/2638549

Yaroslav Ganin, Daniil Kononenko, Diana Sungatullina, and Victor Lempitsky. 2016. DeepWarp: Photorealistic Image Resynthesis for Gaze Manipulation. In *European Conference on Computer Vision (ECCV)*. 311–326. https://doi.org/10.1007/978-3-319-46475-6_20

Pablo Garrido, Levi Valgaerts, Ole Rehmsen, Thorsten Thormaehlen, Patrick Pérez, and Christian Theobalt. 2014. Automatic Face Reenactment. In *Conference on Computer Vision and Pattern Recognition (CVPR)*. 4217–4224. https://doi.org/10.1109/CVPR.2014.537

Pablo Garrido, Levi Valgaerts, Hamid Sarmadi, Ingmar Steiner, Kiran Varanasi, Patrick Pérez, and Christian Theobalt. 2015. VDub: Modifying Face Video of Actors for Plausible Visual Alignment to a Dubbed Audio Track. *Computer Graphics Forum (Eurographics)* 34, 2 (May 2015), 193–204. https://doi.org/10.1111/cgf.12552

Pablo Garrido, Michael Zollhöfer, Dan Casas, Levi Valgaerts, Kiran Varanasi, Patrick Pérez, and Christian Theobalt. 2016. Reconstruction of Personalized 3D Face Rigs from Monocular Video. *ACM Transactions on Graphics* 35, 3 (June 2016), 28:1–15. https://doi.org/10.1145/2890493

Ian J. Goodfellow, Jean Pouget-Abadie, Mehdi Mirza, Bing Xu, David Warde-Farley, Sherjil Ozair, Aaron Courville, and Yoshua Bengio. 2014. Generative Adversarial Nets. In *Advances in Neural Information Processing Systems*.

Geoffrey E. Hinton and Ruslan Salakhutdinov. 2006. Reducing the Dimensionality of Data with Neural Networks. *Science* 313, 5786 (July 2006), 504–507. https://doi.org/10.1126/science.1127647

Liwen Hu, Shunsuke Saito, Lingyu Wei, Koki Nagano, Jaewoo Seo, Jens Fursund, Iman Sadeghi, Carrie Sun, Yen-Chun Chen, and Hao Li. 2017. Avatar Digitization from a Single Image for Real-time Rendering. *ACM Transactions on Graphics (SIGGRAPH Asia)* 36, 6 (November 2017), 195:1–14. https://doi.org/10.1145/3130800.3130887

Alexandru Eugen Ichim, Sofien Bouaziz, and Mark Pauly. 2015. Dynamic 3D Avatar Creation from Hand-held Video Input. *ACM Transactions on Graphics (SIGGRAPH)* 34, 4 (July 2015), 45:1–14. https://doi.org/10.1145/2766943

Phillip Isola, Jun-Yan Zhu, Tinghui Zhou, and Alexei A. Efros. 2017. Image-to-Image Translation with Conditional Adversarial Networks. In *Conference on Computer*

*Vision and Pattern Recognition (CVPR)*. 5967–5976. https://doi.org/10.1109/CVPR.2017.632

Tero Karras, Timo Aila, Samuli Laine, and Jaakko Lehtinen. 2018. Progressive Growing of GANs for Improved Quality, Stability, and Variation. In *International Conference on Learning Representations (ICLR)*.

Ira Kemelmacher-Shlizerman. 2013. Internet-Based Morphable Model. In *International Conference on Computer Vision (ICCV)*. 3256–3263. https://doi.org/10.1109/ICCV.2013.404

Ira Kemelmacher-Shlizerman, Aditya Sankar, Eli Shechtman, and Steven M. Seitz. 2010. Being John Malkovich. In *European Conference on Computer Vision (ECCV)*. 341–353. https://doi.org/10.1007/978-3-642-15549-9_25

Ira Kemelmacher-Shlizerman, Eli Shechtman, Rahul Garg, and Steven M. Seitz. 2011. Exploring photobios. *ACM Transactions on Graphics (SIGGRAPH)* 30, 4 (August 2011), 61:1–10. https://doi.org/10.1145/2010324.1964956

Diederik P. Kingma and Jimmy Ba. 2015. Adam: A Method for Stochastic Optimization. In *International Conference on Learning Representations (ICLR)*.

Christoph Lassner, Gerard Pons-Moll, and Peter V. Gehler. 2017. A Generative Model of People in Clothing. In *International Conference on Computer Vision (ICCV)*. 853–862. https://doi.org/10.1109/ICCV.2017.98

Hao Li, Laura Trutoiu, Kyle Olszewski, Lingyu Wei, Tristan Trutna, Pei-Lun Hsieh, Aaron Nicholls, and Chongyang Ma. 2015. Facial Performance Sensing Head-mounted Display. *ACM Transactions on Graphics (SIGGRAPH)* 34, 4 (July 2015), 47:1–9. https://doi.org/10.1145/2766939

Kai Li, Qionghai Dai, Ruiping Wang, Yebin Liu, Feng Xu, and Jue Wang. 2014. A Data-Driven Approach for Facial Expression Retargeting in Video. *IEEE Transactions on Multimedia* 16, 2 (February 2014), 299–310. https://doi.org/10.1109/TMM.2013.2293064

Kang Liu and Joern Ostermann. 2011. Realistic facial expression synthesis for an image-based talking head. In *International Conference on Multimedia and Expo (ICME)*. https://doi.org/10.1109/ICME.2011.6011835

Ming-Yu Liu, Thomas Breuel, and Jan Kautz. 2017. Unsupervised Image-to-Image Translation Networks. In *Advances in Neural Information Processing Systems*.

Zicheng Liu, Ying Shan, and Zhengyou Zhang. 2001. Expressive Expression Mapping with Ratio Images. In *Annual Conference on Computer Graphics and Interactive Techniques (SIGGRAPH)*. 271–276. https://doi.org/10.1145/383259.383289

Liqian Ma, Qianru Sun, Xu Jia, Bernt Schiele, Tinne Tuytelaars, and Luc Van Gool. 2017. Pose Guided Person Image Generation. In *Advances in Neural Information Processing Systems*.

Mehdi Mirza and Simon Osindero. 2014. Conditional Generative Adversarial Nets. (2014). https://arxiv.org/abs/1411.1784 arXiv:1411.1784.

Kyle Olszewski, Zimo Li, Chao Yang, Yi Zhou, Ronald Yu, Zeng Huang, Sitao Xiang, Shunsuke Saito, Pushmeet Kohli, and Hao Li. 2017. Realistic Dynamic Facial Textures from a Single Image using GANs. In *International Conference on Computer Vision (ICCV)*. 5439–5448. https://doi.org/10.1109/ICCV.2017.580

Kyle Olszewski, Joseph J. Lim, Shunsuke Saito, and Hao Li. 2016. High-fidelity Facial and Speech Animation for VR HMDs. *ACM Transactions on Graphics (SIGGRAPH Asia)* 35, 6 (November 2016), 221:1–14. https://doi.org/10.1145/2980179.2980252

Alec Radford, Luke Metz, and Soumith Chintala. 2016. Unsupervised Representation Learning with Deep Convolutional Generative Adversarial Networks. In *International Conference on Learning Representations (ICLR)*.

Ravi Ramamoorthi and Pat Hanrahan. 2001. An efficient representation for irradiance environment maps. In *Annual Conference on Computer Graphics and Interactive Techniques (SIGGRAPH)*. 497–500. https://doi.org/10.1145/383259.383317

Elad Richardson, Matan Sela, and Ron Kimmel. 2016. 3D Face Reconstruction by Learning from Synthetic Data. In *International Conference on 3D Vision (3DV)*. 460–469. https://doi.org/10.1109/3DV.2016.56

Elad Richardson, Matan Sela, Roy Or-El, and Ron Kimmel. 2017. Learning Detailed Face Reconstruction from a Single Image. In *Conference on Computer Vision and Pattern Recognition (CVPR)*. 5553–5562. https://doi.org/10.1109/CVPR.2017.589

Olaf Ronneberger, Philipp Fischer, and Thomas Brox. 2015. U-Net: Convolutional Networks for Biomedical Image Segmentation. In *International Conference on Medical Image Computing and Computer-Assisted Intervention (MICCAI)*. 234–241. https://doi.org/10.1007/978-3-319-24574-4_28

Joseph Roth, Yiying Tong Tong, and Xiaoming Liu. 2017. Adaptive 3D Face Reconstruction from Unconstrained Photo Collections. *IEEE Transactions on Pattern Analysis and Machine Intelligence* 39, 11 (November 2017), 2127–2141. https://doi.org/10.1109/TPAMI.2016.2636829

Jason M. Saragih, Simon Lucey, and Jeffrey F. Cohn. 2011. Real-time avatar animation from a single image. In *International Conference on Automatic Face and Gesture Recognition (FG)*. 117–124. https://doi.org/10.1109/FG.2011.5771383

Matan Sela, Elad Richardson, and Ron Kimmel. 2017. Unrestricted Facial Geometry Reconstruction Using Image-to-Image Translation. In *International Conference on Computer Vision (ICCV)*. 1585–1594. https://doi.org/10.1109/ICCV.2017.175

Fuhao Shi, Hsiang-Tao Wu, Xin Tong, and Jinxiang Chai. 2014. Automatic Acquisition of High-fidelity Facial Performances Using Monocular Videos. *ACM Transactions on Graphics (SIGGRAPH Asia)* 33, 6 (November 2014), 222:1–13. https://doi.org/10.1145/2661229.2661290

Robert W. Sumner and Jovan Popović. 2004. Deformation Transfer for Triangle Meshes. *ACM Transactions on Graphics (SIGGRAPH)* 23, 3 (August 2004), 399–405. https://doi.org/10.1145/1015706.1015736

Supasorn Suwajanakorn, Ira Kemelmacher-Shlizerman, and Steven M. Seitz. 2014. Total Moving Face Reconstruction. In *European Conference on Computer Vision (ECCV) (Lecture Notes in Computer Science)*, Vol. 8692. 796–812. https://doi.org/10.1007/978-3-319-10593-2_52

Supasorn Suwajanakorn, Steven M. Seitz, and Ira Kemelmacher-Shlizerman. 2015. What Makes Tom Hanks Look Like Tom Hanks. In *International Conference on Computer Vision (ICCV)*. 3952–3960. https://doi.org/10.1109/ICCV.2015.450

Supasorn Suwajanakorn, Steven M. Seitz, and Ira Kemelmacher-Shlizerman. 2017. Synthesizing Obama: Learning Lip Sync from Audio. *ACM Transactions on Graphics (SIGGRAPH)* 36, 4 (July 2017), 95:1–13. https://doi.org/10.1145/3072959.3073640

Yaniv Taigman, Adam Polyak, and Lior Wolf. 2017. Unsupervised Cross-Domain Image Generation. In *International Conference on Learning Representations (ICLR)*.

Ayush Tewari, Michael Zollhöfer, Hyeongwoo Kim, Pablo Garrido, Florian Bernard, Patrick Pérez, and Christian Theobalt. 2017. MoFA: Model-based Deep Convolutional Face Autoencoder for Unsupervised Monocular Reconstruction. In *International Conference on Computer Vision (ICCV)*. 3735–3744. https://doi.org/10.1109/ICCV.2017.401

Justus Thies, Michael Zollhöfer, Matthias Nießner, Levi Valgaerts, Marc Stamminger, and Christian Theobalt. 2015. Real-time Expression Transfer for Facial Reenactment. *ACM Transactions on Graphics (SIGGRAPH Asia)* 34, 6 (November 2015), 183:1–14. https://doi.org/10.1145/2816795.2818056

Justus Thies, Michael Zollhöfer, Marc Stamminger, Christian Theobalt, and Matthias Nießner. 2016. Face2Face: Real-Time Face Capture and Reenactment of RGB Videos. In *Conference on Computer Vision and Pattern Recognition (CVPR)*. 2387–2395. https://doi.org/10.1109/CVPR.2016.262

Justus Thies, Michael Zollhöfer, Marc Stamminger, Christian Theobalt, and Matthias Nießner. 2018. FaceVR: Real-Time Facial Reenactment and Eye Gaze Control in Virtual Reality. *ACM Transactions on Graphics* (2018).

Anh Tuan Tran, Tal Hassner, Iacopo Masi, and Gerard Medioni. 2017. Regressing Robust and Discriminative 3D Morphable Models with a very Deep Neural Network. In *Conference on Computer Vision and Pattern Recognition (CVPR)*. 1493–1502. https://doi.org/10.1109/CVPR.2017.163

Daniel Vlasic, Matthew Brand, Hanspeter Pfister, and Jovan Popović. 2005. Face Transfer with Multilinear Models. *ACM Transactions on Graphics (SIGGRAPH)* 24, 3 (July 2005), 426–433. https://doi.org/10.1145/1073204.1073209

Chao Wang, Haiyong Zheng, Zhibin Yu, Ziqiang Zheng, Zhaorui Gu, and Bing Zheng. 2017. Discriminative Region Proposal Adversarial Networks for High-Quality Image-to-Image Translation. (2017). https://arxiv.org/abs/1711.09554 arXiv:1711.09554.

Ting-Chun Wang, Ming-Yu Liu, Jun-Yan Zhu, Andrew Tao, Jan Kautz, and Bryan Catanzaro. 2018. High-Resolution Image Synthesis and Semantic Manipulation with Conditional GANs. In *Conference on Computer Vision and Pattern Recognition (CVPR)*.

Thibaut Weise, Sofien Bouaziz, Hao Li, and Mark Pauly. 2011. Realtime Performance-based Facial Animation. *ACM Transactions on Graphics (SIGGRAPH)* 30, 4 (July 2011), 77:1–10. https://doi.org/10.1145/2010324.1964972

Erroll Wood, Tadas Baltrušaitis, Louis-Philippe Morency, Peter Robinson, and Andreas Bulling. 2018. GazeDirector: Fully articulated eye gaze redirection in video. *Computer Graphics Forum (Eurographics)* 37, 2 (2018). https://doi.org/10.1111/cgf.13355

Chenglei Wu, Derek Bradley, Markus Gross, and Thabo Beeler. 2016. An Anatomically-Constrained Local Deformation Model for Monocular Face Capture. *ACM Transactions on Graphics (SIGGRAPH)* 35, 4 (July 2016), 115:1–12. https://doi.org/10.1145/2897824.2925882

Zili Yi, Hao Zhang, Ping Tan, and Minglun Gong. 2017. DualGAN: Unsupervised Dual Learning for Image-to-Image Translation. In *International Conference on Computer Vision (ICCV)*. 2868–2876. https://doi.org/10.1109/ICCV.2017.310

Jun-Yan Zhu, Taesung Park, Phillip Isola, and Alexei A. Efros. 2017. Unpaired Image-to-Image Translation using Cycle-Consistent Adversarial Networks. In *International Conference on Computer Vision (ICCV)*. 2242–2251. https://doi.org/10.1109/ICCV.2017.244

Michael Zollhöfer, Justus Thies, Pablo Garrido, Derek Bradley, Thabo Beeler, Patrick Pérez, Marc Stamminger, Matthias Nießner, and Christian Theobalt. 2018. State of the Art on Monocular 3D Face Reconstruction, Tracking, and Applications. *Computer Graphics Forum* 37, 2 (2018). https://doi.org/10.1111/cgf.13382